\documentclass[10pt,journal,compsoc]{IEEEtran}
\usepackage{bm}
\usepackage{url}
\usepackage{xcolor}
\usepackage{booktabs}
\usepackage{multirow}
\usepackage{enumitem}
\usepackage{hyperref}
\usepackage{graphicx}
\usepackage {makecell}
\usepackage{algorithm}

\usepackage{algpseudocode}
\usepackage{threeparttable}
\usepackage{amsmath,amssymb,amsfonts}

\begin{document}

\title{De Novo Generation of Hit-like Molecules from Gene Expression Profiles via Deep Learning}

\author{Chen Li*, Yoshihiro Yamanishi
\IEEEcompsocitemizethanks{\IEEEcompsocthanksitem C. Li and Y. Yamanishi are with the Graduate School of Informatics, Nagoya University, Chikusa, Nagoya, 464-8602, Japan.\protect\\
Email: li.chen.d3c@osaka-u.ac.jp, yamanishi@i.nagoya-u.ac.jp}

\thanks{*. Corresponding author.}}

\markboth{Journal of \LaTeX\ Class Files,~Vol.~14, No.~8, August~2015}{}

\IEEEtitleabstractindextext{
\begin{abstract}
$De~novo$ generation of hit-like molecules is a challenging task in the drug discovery process. Most methods in previous studies learn the semantics and syntax of molecular structures by analyzing molecular graphs or simplified molecular input line entry system (SMILES) strings; however, they do not take into account the drug responses of the biological systems consisting of genes and proteins. In this study we propose a \textbf{\underline{h}}ybrid \textbf{\underline{n}}eural \textbf{\underline{n}}etwork, HNNMol, which utilizes gene expression profiles \textbf{\underline{to}} generate \textbf{\underline{mol}}ecular structures with desirable phenotypes for arbitrary target proteins. In the algorithm, a variational autoencoder is employed as a feature extractor to learn the latent feature distribution of the gene expression profiles. Then, a long short-term memory is leveraged as the chemical generator to produce syntactically valid SMILES strings that satisfy the feature conditions of the gene expression profile extracted by the feature extractor. Experimental results and case studies demonstrate that the proposed HVL2Mol model can produce new molecules with potential bioactivities and drug-like properties.
\end{abstract}

}

\maketitle
\IEEEdisplaynontitleabstractindextext
\IEEEpeerreviewmaketitle

\section{Introduction}
\label{sec:introduction}
Exploring  the chemical space to discover molecules with therapeutic effects (e.g., anticancer drug production) is a time-consuming, costly, and high-risk task in the drug discovery field. Despite extensive premarket drug testing, the failure rate is still $>90\%$ \cite{ding2021identification,kale2023chemograph}. In general, drug development takes over $12$ years and costs greater than $\$1.3$ billion \cite{ertl2003cheminformatics,bongini2021molecular,wouters2020estimated}. After identification of therapeutic target proteins for a disease of interest, researchers search for potential drug candidate molecules that can interact with the therapeutic target proteins. This process is referred to as hit identification \cite{chang2014crowdsourcing,stecula2020discovery}. The high-throughput screening of large-scale chemical compound libraries with various biological assays is often performed for the hit identification, but the experimental approach is quite expensive. 

As an alternative to hit identification, computational methods such as virtual screening \cite{gimeno2019light,melville2009machine} and $de~novo$ molecular generation \cite{schneider2019automated,lin2020review} can be used to accelerate the production of drug candidate molecules. Virtual screening attempts to explore chemical databases containing massive volumes of molecules at minimal cost and obtain hit-like molecules through ligand-based similarity search or docking simulation \cite{shen2012discovery}. $De~novo$ molecular generation attempts to generate new molecules with desired chemical properties or similar to known ligands \cite{mao2021molecular,payne2022bifunctional}. Recently, artificial intelligence and deep learning-based generative models such as variational autoencoders (VAEs) \cite{oliveira2022molecular,dollar2021attention} and generative adversarial networks (GANs) \cite{de2019molgan,litransformer,li2023spotgan} have emerged for the $de~novo$ molecular generation. 
However, most methods in the previous studies focused on learning the syntax and semantics of molecular structures by analyzing molecular graphs or simplified molecular input line entry system (SMILES) strings. 

The biological system is perturbed by drug treatment, thus, the use of biological data in addition to chemical data is desired for drug discovery. Omics data including transcriptome offer a comprehensive molecular landscape that can describe the cellular responses of human cells to drug treatment and the pathological histories of disease patients. Thus, omics data representing drug activities are important resources for current drug development. For example, the use of gene expression data in the preliminary stage of drug discovery is a promising approach \cite{thomas2012impact}, because it does not depend on prior knowledge of ligand structures or three-dimensional (3D) structural information of therapeutic target proteins \cite{krishnan2021novo,bung2022silico}. However, omics-based drug discovery approach has severe limitations. The number of molecules with omics information is quite limited; thus, the method is applicable only to molecules for which omics data are measured. Deep learning-based methods have been developed to generate hit-like molecules from gene expression data using GAN \cite{mendez2020novo} and VAE \cite{kaitoh2021triomphe}, but generated molecules tend to be chemically invalid  or have unrealistic structures, thus, there remain room in terms of accuracy improvement.

In this study, we present a deep generative model, HVL2Mol, to analyze omics data and design new drug structures to overcome the above problems. Specifically, a VAE model is utilized as a feature extractor to learn the low-dimensional features of the gene expression profile data. Then, a long short-term memory (LSTM) \cite{li2018capturing,zhang2019multi,sagheer2019time} model is leveraged as the chemical generator to produce syntactically valid SMILES strings that satisfy the feature conditions of the gene expression profile extracted by the VAE-based feature extractor. The features of the gene expression profiles are involved throughout the training process as conditions for the LSTM model, which can guide the model to generate molecules associated with the target gene expression profile. Our primary contributions are as follows:
\begin{itemize}[leftmargin=0.3cm]
\item {\bf A novel idea}: unlike the previous methods on the generation of molecular chemical structures (e.g., SMILES strings and graph structures), this study attempts to generate hit-like molecules from scratch using gene expression profiles.
\item {\bf A concise model}: combining simple deep learning models (i.e., VAE and LSTM models) achieves the complex goal of molecular generation considering biological information.
\item {\bf Superior performance}: the experimental results demonstrate that the proposed method yields new molecules with potential bioactivities and drug-likeness properties, which can be utilized for further structure optimization.
\end{itemize}

The rest of the study is organized as follows. We survey existing work related to various deep learning-based molecular generation methods in section \ref{sec:related}. Section \ref{sec:model} presents the proposed HVL2Mol model for $de~novo$ molecular generation using gene expression profiles. In section \ref{sec:exp}, we conduct comprehensive experiments and case studies to compare with the SOTA models to validate the effectiveness of the proposed HVL2Mol model. Finally, we summarize the paper and show the future direction in section \ref{sec:conclusion}.

\section{Related Work}
\label{sec:related}
In this section, our emphasis is on conducting a comprehensive review of previous studies encompassing various molecular generation approaches. This includes conventional methods, deep learning-based $de~novo$ methods, and omics data-driven hit-like molecule generation.

\subsection{Conventional Molecular Generation Methods}
\label{sec:conventional}
In the realm of molecular generation for drug discovery, conventional methods have long relied on chemical intuition \cite{wang2022theory}, medicinal chemistry principles \cite{wermuth2011practice}, combinatorial chemistry, and structure-based design \cite{akaji2011structure}. Experienced chemists use their knowledge to design molecules, modify existing structures based on medicinal chemistry principles, synthesize diverse compound libraries through combinatorial chemistry, and leverage structural information for targeted design.

However, these conventional methods exhibit limitations. Human bias and intuition, inherent in chemical design, may restrict the exploration of vast chemical space. Traditional approaches are time-consuming, expensive, and may not efficiently explore diverse molecular structures for high-throughput screening \cite{ding2021identification}. Predicting bioactivity based solely on chemical intuition can be challenging, and conventional methods struggle to capture complex relationships between molecular structures and biological activities \cite{kale2023chemograph}.

\subsection{Deep Learning-based Molecular Generation}
\label{sec:deep}
To tackle these challenges, deep learning-based $de~novo$ molecular generation, a relatively recent field of research, integrates the capabilities of machine learning and high-throughput data analysis techniques. Its primary objective is to generate new molecules with desired bioactivities, utilizing molecular graphs, Self-referencing embedded strings (SELFIES) \cite{krenn2020self}, and SMILES strings as the primary data types in drug design processes \cite{jin2018junction,du2022interpretable,du2022small,zang2020moflow,frey2022fastflows,vignac2023digress}. These methods offer the potential to explore broader regions of chemical space, predict bioactivity more accurately, and expedite the drug discovery process.

\subsubsection{Graph-based Methods}
\label{sec:graph}
Molecular graphs contain rich structural information and are often used for molecular generation and drug design \cite{jin2018junction}. Typically, a molecular graph is usually represented by an ensemble of atom vectors and bond matrices. VAE models attempts to approximate the distribution of molecular graphs to learn latent variables \cite{du2022interpretable,du2022small}. 

Generally, VAE-based models construct molecular graphs with a tree structure and employ an encoder to extract the molecular graph features and represent them as low-dimensional latent vectors. Then, the VAE decoder is employed as a molecular generator to reconstruct atoms in the tree into molecules via the latent vector representation. The design of graph-based generators is challenging; thus, GAN-based molecular generation models are rare. The molecular GAN (MolGAN) method \cite{de2019molgan} generates new graphs with the maximum likelihood of atoms and chemical bonds by sampling atomic features and chemical bond feature matrices. In addition, an actor-critic \cite{lillicrap2015continuous} reward network is used to calculate the property scores of the generated graphs. However, MolGAN suffers from a severe mode collapse, thereby causing its uniqueness to be less than 5\%. ALMGIG \cite{polsterl2021adversarial}, which is an extension of the bidirectional GAN model that generates new molecules by learning distributions in the molecular space using adversarial cyclic consistency loss.

Flow-based molecular generative models, exemplified by MoFlow \cite{zang2020moflow}, initially produce bonds (edges) using a Glow-based model. Subsequently, atoms (nodes) are generated based on the established bonds through a novel graph conditional flow. Finally, these components are assembled into a chemically valid molecular graph, with posthoc validity correction. Diffusion-based molecular generative models, such as DiGress \cite{vignac2023digress}, are based on a discrete diffusion process. Graphs are iteratively modified with noise through the addition or removal of edges and changes in categories.

\subsubsection{SMILES-based Methods}
\label{sec:smiles}
$De~novo$ drug design using SMILES strings attempts to generate new molecules with desired properties \cite{kusner2017grammar,litransformer,song2023dnmg}. For example, GrammarVAE \cite{kusner2017grammar} is a SMILES-based model that is used to generate molecular structures, where a VAE is used with a grammar-based decoder that generates syntactically valid SMILES strings. This model is trained on a dataset of existing molecules and generates new molecules with high structural diversity. In addition, TransORGAN \cite{litransformer} is a transformer-based GAN model designed to generate diverse molecules that are similar to the source molecules. In the TransORGAN model, the transformer architecture and a one-dimensional convolutional neural network are employed as the generator and discriminator, respectively, and the Monte Carlo policy gradient algorithm \cite{silver2009monte} is used to explore new molecules with high chemical properties. DNMG \cite{song2023dnmg} is a transfer learning-based GAN that considers the 3D grid space information of the ligand with atomic physicochemical properties in the molecular generation process. Then, the generated molecules are parsed into SMILES strings using a captioning network.

\subsubsection{SELFIES-based Methods}
\label{sec:selfies}
Introduced more recently, SELFIES \cite{krenn2020self} tackles the challenge of string invalidity at a profound level by purportedly offering representations for all molecules, thereby guaranteeing the validity of every SELFIES string. Each symbol within a SELFIES string is derived from a rule vector that precisely denotes the chemical structure type, such as [C] for carbon atoms or [=O] for double-bonded oxygen, while the state of derivation encapsulates both syntactic and chemical constraints, including considerations like maximal valency \cite{wigh2022review}. This robust framework ensures the integrity and reliability of SELFIES representations. This makes SELFIES particularly suitable for machine learning applications in chemistry. The flexibility of SELFIES enables the encoding of complex molecular structures. FastFlows \cite{frey2022fastflows} utilizes normalizing-flow based models, SELFIES, and multi-objective optimization to generate small molecules. Notably, it can produce thousands of chemically valid molecules within seconds, even with an initial training set as small as 100 molecules.

While SELFIES has showcased remarkable efficacy in molecular representation, and a comprehensive library\footnote{\url{https://github.com/aspuru-guzik-group/selfies}} is available for the translation between SMILES strings and SELFIES representations \cite{krenn2022selfies}, it nonetheless exhibits certain limitations. Firstly, the representation of SELFIES is characterized by a relatively high level of complexity and low interpretability, especially when dealing with advanced molecular grammar. This complexity and interpretability often poses challenges in accurately deciphering certain sequences \cite{wu2024tsis}. In contrast, SMILES strings offer advantages in widespread adoption, conciseness, and ease of readability. Secondly, unlike SMILES strings, which focus on encoding the semantics of molecules, SELFIES is primarily designed to generate strings representing syntactically valid molecular structures \cite{mukherjee2021predicting,das2024advances}. The intricate semantics of SELFIES entail that even subtle adjustments in syntax can lead to the generation of molecular structures that are significantly divergent in terms of their chemical composition, functional groups, and overall properties \cite{vogt2023exploring}. Also, SELFIES faces limitations in fully representing certain macromolecules and crystals, particularly those comprised of large molecules or characterized by intricate bonding patterns \cite{krenn2022selfies}.

Based on the above considerations, SMILES strings maintain their position as a reliable option for molecular representation, providing a robust framework for various cheminformatics applications. Thus, in this study, we employ SMILES strings as the choice for molecular generation.

\subsection{Omics Data-driven Molecule Generation}
\label{sec:omics}
To date, most methods in previous studies generated hit-like molecules based on a learning set of ligand structures and bioactivities, where the structures are represented by graphs or SMILES strings. Diverging from conventional approaches, omics data-driven hit-like molecule generation endeavors to leverage omics data, specifically gene expression profiles. The overarching goal is to generate hit molecules that exhibit promising biological activity against specific targets, such as proteins or enzymes associated with particular diseases. To the best of our knowledge, there are limited studies that have explored drug design directly from omics data \cite{mendez2020novo,kaitoh2021triomphe}.

Generally, omics-based methods can generate hit-like molecules without prior knowledge of ligand structures and the 3D structure of the target proteins. A conditional Wasserstein GAN combined with a gradient penalty was proposed to generate hit-like molecules from noise using gene expression profile data \cite{mendez2020novo}, which is referred to as ConGAN in this study. However, the validity of the generated candidate molecules is not guaranteed, thereby limiting the hit identification ability. In addition, the prediction process of transcriptional correlation between ligands and targets is unclear. TRIOMPHE \cite{kaitoh2021triomphe} is a VAE-based molecular generation model using transcriptional correlation between the gene expression profile with the perturbation of a therapeutic target protein and the gene expression profile with the treatment of small molecules. The most similar molecule is selected as the source molecule, the source molecule is projected to the latent space using a VAE encoder, and a decoder is used to sample and decode the latent vectors into new molecules. However, in their work, gene expression profiles were solely employed in correlation calculations for selecting SMILES strings before inputting them into the VAE model. During the molecular generation phase, gene expression profiles were not utilized to guide the generation of hit-like molecules. Consequently, the molecules generated using TRIOMPHE exhibited low Tanimoto coefficients compared to the corresponding known ligands.

Note that, unlike the previously mentioned $de~novo$ molecular generative models, the proposed HVL2Mol aims to generate hit-like molecules that exhibit promising biological activity against specific target proteins or particular diseases, utilizing information from gene expression profiles. HVL2Mol initially extracts biological features from gene expression profiles using a VAE model. Subsequently, these extracted features serve as conditions for the conditional LSTM model, guiding the generation of hit-like molecules.

\section{Proposed Model}
\label{sec:model}

\begin{figure*}[t]
\centering
\includegraphics[width=0.85\textwidth]{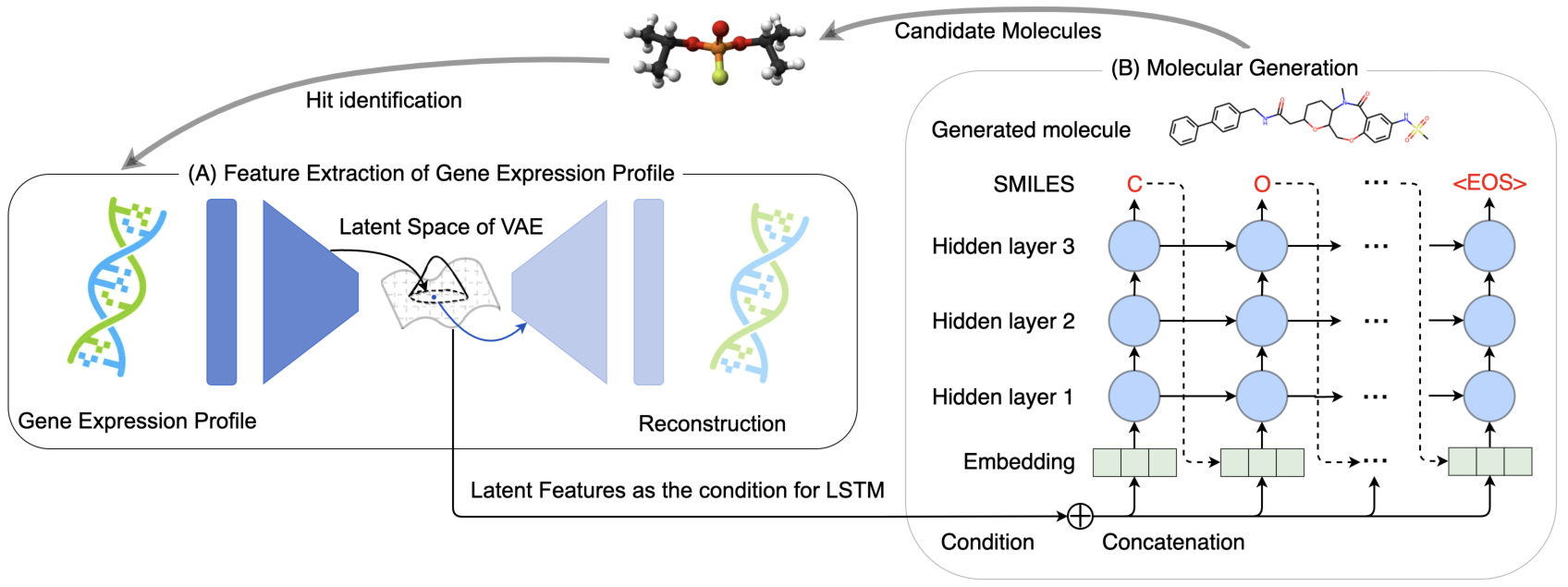}
\vspace{-10pt}
\caption{Architecture of the proposed HVL2Mol model. (A) A VAE is trained to extract the biological features of gene expression profiles. Here, a VAE encoder attempts to extract the latent feature vector of a gene expression profile, and a VAE decoder attempts to reconstruct the gene expression profile from the latent vector. (B) After the VAE training, the latent vector is utilized as a condition to an LSTM to generate SMILES strings. An extracted latent vector and a vector representation of a start token are concatenated to generate the first atom of a SMILES string. Then, the generated atom and the condition generate the next atom iteratively. This iterative process ends when the defined end token (i.e., $<$EOS$>$) is generated. Finally, all atoms are combined to form a SMILES string. The newly generated SMILES string can be used as a candidate molecule for hit identification to treat diseases.}
\label{fig:gx2mol}
\end{figure*}

\subsection{HVL2Mol}
\label{sec:gx2mol}
The proposed HVL2Mol comprises two main components. The feature extractor (i.e., the VAE model) is employed to extract the features of the gene expression profiles, and the generator (i.e., the conditional LSTM model) is used to generate hit-like molecules based on the extracted features of gene expression profiles. 

We aim to generate hit-like molecules from gene expression profiles. High-dimensional gene expression profiles present formidable challenges attributed to the presence of noise and redundant information. Employing a VAE model for feature extraction and selection from such intricate datasets emerges as a strategic solution. VAEs, belonging to the category of generative models, possess the capability to comprehend the complexities of high-dimensional data by acquiring a lower-dimensional representation while encapsulating its inherent structure. This method serves as an effective means to navigate and mitigate the issues associated with noise and redundancy in gene expression profiles. The subsequent elucidation highlights the versatile application of VAEs in overcoming these challenges.
\begin{itemize}[leftmargin=0.3cm]
\item \textbf{Dimensionality reduction:} VAEs can effectively reduce the dimensionality of high-dimensional gene expression profiles by learning a compressed and meaningful representation in the latent space. The encoder part of the VAE maps the input gene expression profiles to a lower-dimensional latent space, capturing essential features.

\item \textbf{Noise reduction:} VAEs are designed to model the underlying distribution of the data. This can help in filtering out noise and capturing the intrinsic patterns present in the gene expression profiles. The generative nature of VAEs encourages the model to focus on the most salient features while ignoring noise.

\item \textbf{Feature extraction:} The latent space learned by the VAE can be considered as a set of extracted features that represent the essential information in the gene expression profiles. These features can serve as a more compact and informative representation of the data compared to the original high-dimensional space.
\end{itemize}

\subsection{Extraction of Biological Features} 
\label{se:extraction}
The architecture of the HVL2Mol model is illustrated in Figure \ref{fig:gx2mol}. In phase (A), we initiate the process by training a VAE model, extracting essential biological features from gene expression profiles. The encoder network transforms the features of gene expression profiles into a low-dimensional latent space, which is subsequently reconstructed by the decoder. Post the training phase, only the encoder is utilized for subsequent downstream tasks.

Formally, let $\bm{G}=[g_1, g_2,\cdots, g_T]$ indicate the gene expression profile, where $g_i$ represents the $i$-th gene with the maximum gene number of $T$. The VAE model serves as a feature extractor in our HVL2Mol model, tasked with learning a latent feature distribution denoted as $p(z|\bm{G})$. The objective is to align this distribution as closely as possible to the reference distribution $p(z)$, characterized as an isotropic normal distribution. This alignment occurs through the approximation of observed gene expression profiles, while reinforcing the stochastic independence among latent variables. The utilization of the VAE model in this manner facilitates the extraction of meaningful latent features from the input data, as demonstrated in Figure \ref{fig:gx2mol} (A). This visualization provides a tangible representation of how the VAE contributes to capturing essential characteristics within the gene expression profiles. High-dimensional gene expression profile reconstruction can be modeled by the integration of the low-dimensional feature space $p(z)$ and conditional distribution $p_{\bm \theta}(\bm{G}|z)$ parameterized by ${\bm \theta}$:
\begin{align}
p_{\bm\theta}(\bm{G}) = \int p_{\bm\theta}(\bm{G}|z)p(z)dz.
\end{align}
To address the intractable issue of the posterior distribution $p_{\bm\theta}(z|\bm{G})$, the feature extractor replaces $p_{\bm\theta}(z|\bm{G})$ by an approximate variational distribution $q_{\bm\theta^{\prime}}(z|\bm{G})$. Typically, $q_{\bm\theta^{\prime}}(z|\bm{G})$ and $p_{\bm\theta}(\bm{G}|z)$ are used as the encoder and decoder of a VAE, respectively. According to the evidence lower bound \cite{ramapuram2020lifelong}, the loss function of the feature extractor can be formulated as follows:
\begin{align}
\label{eq:vae_loss}
\mathcal{L}_{F}({\bm\theta, \bm\theta^{\prime}})=&-\mathbb{E}_{z \sim q_{\bm\theta^{\prime}(z|\bm{G})}}[\log p_{\bm\theta}(\bm{G}|z)]\\\nonumber
&+\beta \cdot D_{KL}(q_{\bm\theta^{\prime}}(z|\bm{G})||p(z)),
\end{align}
where $\mathbb{E}[\cdot]$ and $\beta$ indicate an expectation operation and the weight of the Kullback-Leibler divergence $D_{KL}$ \cite{joyce2011kullback}, respectively. The VAE encoder generates both a mean ($\mu$) and a variance ($\sigma^2$) for each point in the latent space, typically following a Gaussian distribution. For a given gene expression profile $\bm{G}$, calculated as follows:
\begin{equation}
q_{\bm\theta^{\prime}}(z|\bm{G}) = \mathbf{N}\left(\mu(\bm{G}), \sigma^2(\bm{G})\right),
\end{equation}
where $\mu(\bm{G})$ and $\sigma^2(\bm{G})$ are the mean and variance functions parameterized by the encoder. The VAE then samples a point $z$ from this distribution. Finally, the extracted latent vector of the gene expression profiles is as follows: 
\begin{align}
\label{eq:vae_encoder}
\bm{F}_{Gx}=\text{Encoder}(\bm{G}). 
\end{align} 

\subsection{Generation of Hit-like Molecules} 
\label{sec:hit}
Here, an LSTM model is used as the chemical generator to produce syntactically valid SMILES strings that satisfy the feature conditions of the gene expression profiles extracted by the feature extractor. During phase (B), we incorporate the corresponding SMILES strings as inputs for LSTM training. The extracted features from gene expression profiles are fused with each SMILES token, serving as input for the model to iteratively generate the subsequent token in the SMILES string.

Formally, let $\bm{X}_{1:n}=[\bm{x}_1, \bm{x}_2,\cdots, \bm{x}_n]$ denote a SMILES string of length $n$, where $\bm{x}_i$ is the $i$-th embedding vector of the SMILES string with the size of $M$. Then, $\bm{x}_i$ is concatenated with $\bm{F}_{Gx}$ as the input to the generator. The generator iteratively generates a character of the SMILES string at the current time step based on the previous time step. Let $\bm{Y}_{1:n}=[\bm{y}_1, \cdots, \bm{y}_n]$ indicate the predicted SMILES string for $\bm{X}_{1:n}$. According to the negative log likelihood, the loss function of the generator can be calculated as follows:
\begin{align}
\label{eq:lstm_loss}
\mathcal{L}_{G}(\bm{X}_{1:n}, \bm{Y}_{1:n})= -\sum_{i=1}^{n}\log p(\bm{y}_i|\bm{X}_{1:i-1}; \bm{\phi}),
\end{align}
where $\bm{\phi}$ is the parameters of the chemical generator.

During the generation phase, the input to the VAE encoder exclusively comprises gene expression profiles for feature extraction. The resulting extracted features are subsequently employed to steer the process of generating hit-like molecules. Algorithm \ref{alg:gx2mol} summarizes the procedure of the proposed HVL2Mol model. Here, sets of the gene expression profiles and SMILES strings are first used to train the feature extractor and chemical generator. In the training phase, the features of gene expression profiles are learned from a VAE-based feature extractor. The extracted features are used as conditions of the LSTM-based molecular generator. In the testing phase, the gene expression profile is employed to generate new hit-like molecules. 

\begin{algorithm}[t]
\caption{Procedure for the HVL2Mol model}
\label{alg:gx2mol}
\begin{algorithmic}[1]
\State{\bfseries Data:} Gene expression profiles $\bm{G}$ and SMILES strings $\bm{X}_{1:n}$
\State{\bfseries Initialization:} the feature extractor $F_{\theta}$, the molecule generator $G_{\phi}$
\State // Train the feature extractor.
\For{$i=1 \to f\_epochs$}
\State Update $F_{\theta}$ using $\bm{G}$ according to the loss function of Eq. (\ref{eq:vae_loss}).
\EndFor

\State // Train the molecule generator.
\For{$i=1 \to g\_epochs$}
\State Update $G_{\phi}$ using $\bm{X}_{1:n}$ according to the loss function of Eq. (\ref{eq:lstm_loss}).
\EndFor

\State // Generate hit-like molecules from scratch.
\State Extract the features $\bm{F}_{Gx}$ using $\bm{G}$ according to Eq. (\ref{eq:vae_encoder}).
\State Generate the corresponding SMILES representation from $\bm{F}_{Gx}$.

\State // Test the generation task.
\State Calculate the Tanimoto coefficient using known ligands.
\State Select the molecule with the maximum Tanimoto coefficient score as the candidate molecule.
\end{algorithmic}
\end{algorithm}

\section{Experiments}
\label{sec:exp}

\subsection{Experimental Setup}
\label{sec:setup}

\noindent{\bf Datasets.} In this study, we used chemically induced gene expression profiles as training data to train the proposed HVL2Mol model. In addition, we analyzed eight knockdown and two overexpressed target protein perturbed expression profiles to generate hit-like molecules, and disease reversal gene expression profiles as a case study to generate therapeutic molecules.

\begin{itemize}[leftmargin=0.3cm]
\item \textbf{Chemically-induced gene expression profiles} were collected from the Library of Integrated Network-based Cellular Signatures (LINCS) database \cite{duan2014lincs}. The LINCS database stores the gene expression profiles with a dimension of 978 for 77 human cultured cell lines exposed to various molecules. Here we analyzed the gene expression profiles of the MCF7 cell line treated with 13,755 molecules whose SMILES string lengths were less than 80 at a concentration of 10 {\textmu}M.

\item \textbf{Target-perturbed gene expression profiles} were collected from the LINCS database. Here, we analyzed the RAC-alpha serine / threonine-protein kinase (AKT1), RAC-beta serine / threonine-protein kinase (AKT2), Aurora B kinase (AURKB), cysteine synthase A (CTSK), epidermal growth factor receptor (EGFR), histone deacetylase 1 (HDAC1), mammalian target of rapamycin (MTOR), phosphatidylinositol 3-kinase catalytic subunit (PIK3CA), decapentaplegic homologue 3 (SMAD3), and tumor protein p53 (TP53), which have been verified to be useful therapeutic target proteins against cancers. The gene expression profiles for the first eight proteins were obtained from gene knockdown profiles of the MCF7 cell line, while those for the latter two proteins were obtained from gene overexpression profiles. When multiple profiles were measured under different experimental conditions for a single protein, we averaged the multiple profiles of the same target protein to create target protein-specific profiles.

\item \textbf{Disease-specific gene expression profiles} were obtained from the crowd extracted expression of differential signatures (CREEDS) database \cite{wang2016extraction}, which contains the expression profiles of 14,804 genes for 79 diseases. The disease-specific gene expression profiles were acquired by averaging the gene expression profiles from multiple patients with the same disease. Here, we extracted the most relevant 884 genes for gastric cancer, atopic dermatitis, and Alzheimer's disease from the disease-specific gene expression for model validation, and we created the disease reversal profiles by multiplying the disease-specific gene expression by -1. Note that the disease reversal profiles of a disease are considered to be associated with a therapeutic effect on that disease.
\end{itemize}

\noindent{\bf Hyperparameters.} For the feature extractor, the encoder of the VAE  included three feedforward layers with dimensions of $512$, $256$, and $128$. The latent vector dimension was set to 64. Note that the dimensions of the decoder were the opposite dimensions of the encoder, i.e., $128$, $256$, and $512$. The dropout probability \cite{srivastava2014dropout} and learning rate were set to $0.2$ and $1\mathrm{e}{-4}$, respectively. The training of gene expression profiles was conducted with a batch size set at 64. For the generator, the embedding size was set to $128$. The LSTM model contained three hidden layers with dimensions of $256$. The dropout probability and learning rate were set to $0.1$ and $5\mathrm{e}{-4}$, respectively. The maximum length of the generated SMILES strings was fixed to $100$. The batch size for training LSTM was set to 64. In addition, the feature extractor and generator used the Adam optimizer \cite{kingma2014adam}, and the number of training epochs for the feature extractor and generator was set to $2000$ and $300$, respectively. Note that all experiments were conducted on GPUs using CUDA\footnote{Our source code is available on GitHub at  \url{https://github.com/naruto7283/Gx2Mol}}.

\noindent{\bf Dataset splitting and model selection.} The dataset was partitioned into distinct sets for training (80\%), validation (10\%), and testing (10\%) to ensure a robust evaluation of our HVL2Mol model. This division allows for effective model training on the training set, tuning of hyperparameters based on the validation set, and unbiased assessment of model performance on the test set. The selection of the optimal model was determined by monitoring the convergence of the total loss function of HVL2Mol during training. Convergence of the loss function indicates stability and optimal performance. This approach ensures the selection of a well-performing model based on its ability to minimize the defined loss and generalize effectively to unseen data.

\subsection{Evaluation Measures}
\label{sec:metrics}
In this study, three statistical indices (validity, uniqueness, and novelty), along with two essential chemical properties (quantitative estimate of drug-likeness (QED) \cite{bickerton2012quantifying} and synthesizability (SA) \cite{ertl2009estimation}), and the Tanimoto coefficient \cite{racz2018life} were employed to assess hit-like molecules generated by the proposed HVL2Mol model.

\begin{itemize}[leftmargin=0.3cm]
\item \textbf{Validity} denotes the ratio of valid molecules to the total number of training SMILES strings. In practice, this measure is typically calculated using the RDKit tool \cite{landrum2013rdkit}. 
\item \textbf{Uniqueness} refers to the proportion of non-repeated molecules within the set of generated valid molecules.
\item \textbf{Novelty} is defined as the ratio of newly generated valid molecules sharing the identical gene expression profiles but exhibiting distinct (canonical) molecular SMILES representations within the training set.
\item \textbf{QED} can be calculated by assigning different weights to eight molecular descriptors (i.e., molecular weight, octanol-water partition coefficient, number of hydrogen bond donors, number of hydrogen bond acceptors, molecular polar surface area, number of rotatable bonds, number of aromatic rings, and number of structural alarms) \cite{kwon2021molfinder,sun2022prediction}.
where $d_i$ and $W_i$ represent the desirability function and weight of the $i$-th descriptor, respectively. Typically, the weights of the eight molecular descriptors were obtained through chemical experiments. In practice, the QED score was calculated by a function in the RDKit tool. The larger the QED score, the more drug-like the molecule.
\item \textbf{Synthesizability (SA)} is assessed through the SA score, denoted as $
\text{SA} = r_s - \sum_{i=1}^5 p_i$. Here, $r_s$ signifies the “synthetic knowledge," representing the ratio of contributions from all fragments to the total number of fragments in the molecule. In this study, $r_s$ is computed from experimental results \cite{ertl2009estimation}. Each $p_i$ ($i\in\{1,\cdots, 5\}$) corresponds to the ring complexity, stereo complexity, macrocycle penalty, size penalty, and bridge penalty, computed using the RDKit tool. A higher SA score indicates greater ease of synthesizing the molecule.

\item \textbf{Tanimoto coefficient}, which is calculated from the ECFP4 fingerprint \cite{rogers2010extended,ortiz2013improving} with a dimension of 2048. In practice, the ECFP4 and Tanimoto coefficients were calculated using the “GetMorganfingerprintAsBitVect" and “BulkTanimotoSimilarity" functions of the RDKit tool.
\end{itemize}

\subsection{HVL2Mol Training}
\label{sec:res}
We evaluated the effectiveness of the VAE model in extracting the biological features from gene expression profiles and the capability of the LSTM model to generate new molecules experimentally. 
\begin{figure}[t]
\centering
\includegraphics[width=0.7\hsize]{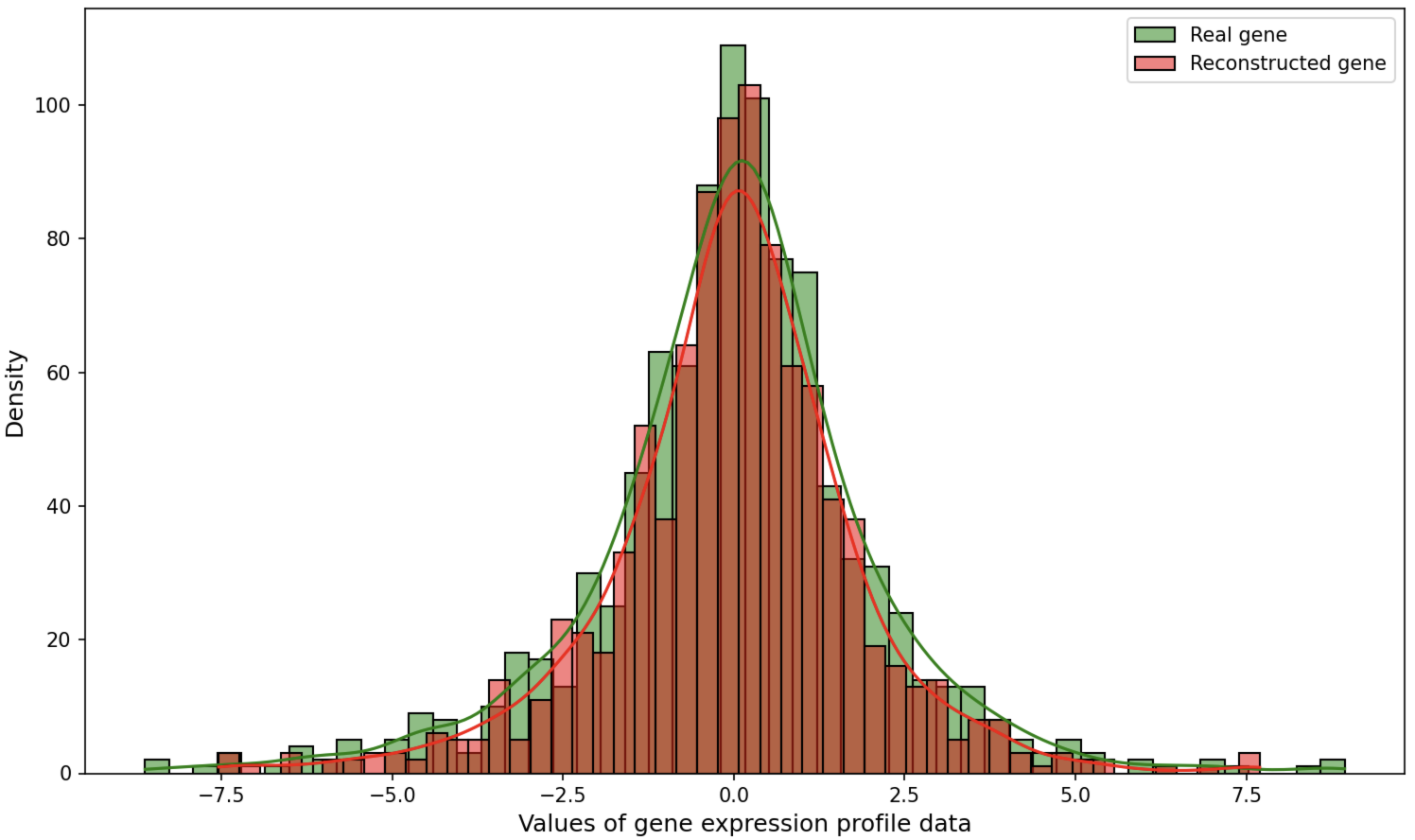}
\vspace{-10pt}
\caption{Distribution of fold change values in the gene expression profile of the molecule “C17H25ClN2O3" exposed in the MCF7 cell. The original gene expression profile of “C17H25ClN2O3" (green) and the reconstructed gene expression profiles (red) have similar distributions.}
\label{fig:one}
\end{figure}

\begin{figure}[ht]
\centering
\includegraphics[width=0.7\hsize]{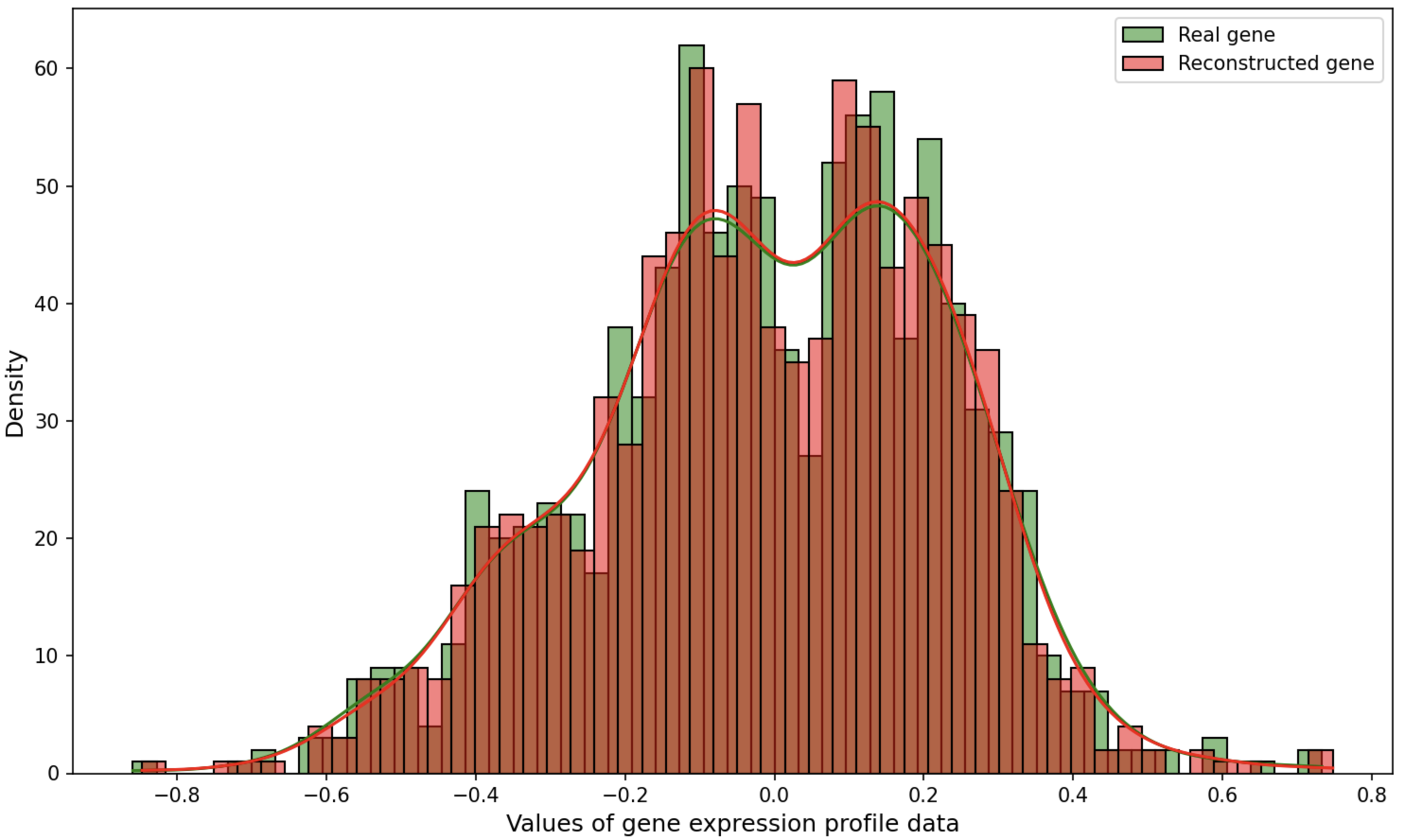}
\vspace{-10pt}
\caption{Distribution of fold change values in the average gene expression profile of all molecules exposed in the MCF7 cell. The original gene expression profiles of the training set (green) and the reconstructed gene expression profiles (red) have similar distributions.}
\label{fig:all}
\end{figure}

Figure \ref{fig:one} shows a comparison of the distribution of fold change values in the gene expression profile of a molecule between the training set and the reconstructed set. Figure \ref{fig:all} shows a comparison of the distribution of fold change values in the average gene expression profile of all molecules between the original set and the reconstructed set. Note that Figure \ref{fig:one} shows the distribution of a gene expression profile of the molecule “C17H25ClN2O3" exposed in the MCF7 cell, whose SMILES representation is denoted as “CCC1=CC(=C(C(=C1O)C(=O)NC[C@@ H]2CCCN2CC)OC)Cl." The distribution of the original gene expression profiles was similar to that of the reconstructed gene expression profiles acquired using the proposed HVL2Mol model. In other words, the VAE utilized in the proposed HVL2Mol model captures the biological features of the gene expression profiles and successfully reconstructs them into the original distribution. 

Note that, in contrast to previous studies like MoFlow \cite{zang2020moflow}, FastFlows \cite{frey2022fastflows}, and DiGress \cite{vignac2023digress}, which focused on generating molecules from scratch, our study is specifically oriented towards generating hit-like molecules from gene expression profiles. Furthermore, due to the limited validity of the ConGAN baseline model, which is below 8.5\% \cite{mendez2020novo}, we restrict our comparison to the proposed HVL2Mol model and the TRIOMPHE baseline model. The statistical results are detailed in Table \ref{tab:statistical_res}, offering a comprehensive comparison with TRIOMPHE. We conducted a comparison between our HVL2Mol model and two variants of the TRIOMPHE model: one based on SELFIES and the other on SMILES-based hit-like molecule generation. Additionally, the LSTM architecture within TRIOMPHE is implemented in two variants, namely unidirectional LSTM (uni-dir) and bidirectional LSTM (bi-dir). In both SELFIES- and SMILES-based molecular generation, the unidirectional LSTM-based TRIOMPHE demonstrated greater uniqueness than its bidirectional LSTM-based model. Furthermore, SMILES-based molecular generation exhibited superior performance compared to SELFIES-based tasks. With the gene expression profiles serving as conditions for HVL2Mol's LSTM, HVL2Mol generated 1322 molecules. The validity, uniqueness, and novelty of the generated molecules using HVL2Mol are 88.6\% (1171 valid molecules), 83.0\% (972 unique molecules), and 99.7\% (1167 novel molecules). Our HVL2Mol showcases a notable improvement of 10.7\% when compared to the best result obtained by the SMILES-based unidirectional TRIOMPHE variant (75.0\%). Overall, HVL2Mol model proves to be effective in the task of hit-like molecular generation.

\begin{table}[t]
\caption{Comparison of key statistical results for the molecules generated by HVL2Mol and baseline models.}
\label{tab:statistical_res}
\centering
\begin{threeparttable}
\begin{tabular}{llccc}
\hline
& Method & Validity & Uniqueness & Novelty \\\hline
\multirow{2}{*}{SELFIES-based} & uni-dir & - & 69.5\% & - \\
& bi-dir & - & 50.2\% & - \\\hline
\multirow{4}{2em}{SMILES-based} & uni-dir & - & 75.0\% & - \\
& bi-dir & - & 70.6\% & - \\\cmidrule{2-5}
& HVL2Mol & 88.6\% & \colorbox[gray]{0.9}{{\bf 83.0\%}} & 99.7\% \\\hline
\end{tabular}
\begin{tablenotes}
\footnotesize
\item[$\star$] 
“uni-dir” and “bi-dir” represent the unidirectional and bidirectional LSTM, respectively, in the TRIOMPHE baseline model.
\end{tablenotes}
\end{threeparttable}
\end{table}

\begin{figure}[t]
\centering
\includegraphics[width=0.7\hsize]{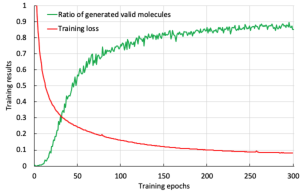}
\vspace{-10pt}
\caption{Training loss and the ratio of valid molecules generated by the proposed HVL2Mol. The red curve indicates the training loss value of the LSTM with the training epochs. The green curve denotes the ratio of valid molecules generated by the LSTM with the training epochs. Note that the valid molecules are examined by the RDKit tool.}
\label{fig:training_res}
\end{figure}

Figure \ref{fig:training_res} shows the training loss and the ratio of the generated valid molecules of the LSTM in the proposed HVL2Mol model. The loss decreased smoothly over the 300 training epochs and finally converges under 0.1. In contrast, the validity of the molecules generated by the conditional LSTM model gradually increased as training proceeds, with the final validity ratio converging at approximately 90\%. Overall, the results indicate that the conditional LSTM model utilized in the proposed HVL2Mol can generate valid molecules effectively.

\begin{figure}[t]
\centering
\includegraphics[width=0.7\hsize]{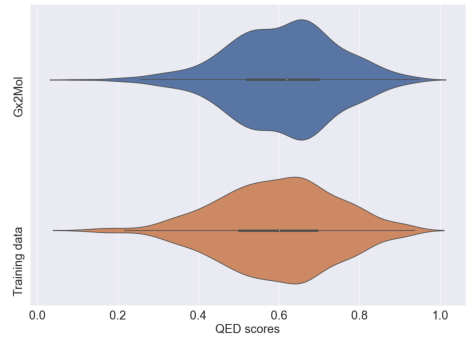}
\vspace{-10pt}
\caption{Violin plots of QED scores for molecules from the training dataset and proposed HVL2Mol.}
\label{fig:violin}
\end{figure}

To further explore the ability of the proposed HVL2Mol model to generate molecules, we also compared the distribution of the QED scores of the molecules generated by HVL2Mol with molecules in the training data. Figure \ref{fig:violin} shows that the generated molecules and the original molecules have similar QED distributions. The average QED scores of molecules in the training dataset and molecules generated by HVL2Mol were 0.60 and 0.61, respectively. The violin plots of the QED scores indicate that the proposed HVL2Mol model did not change the potential chemical property characteristics of the training data during the molecular generation process, which demonstrates the LSTM model's ability to generate molecules effectively. 
\begin{figure}[ht]
\centering
\includegraphics[width=0.7\hsize]{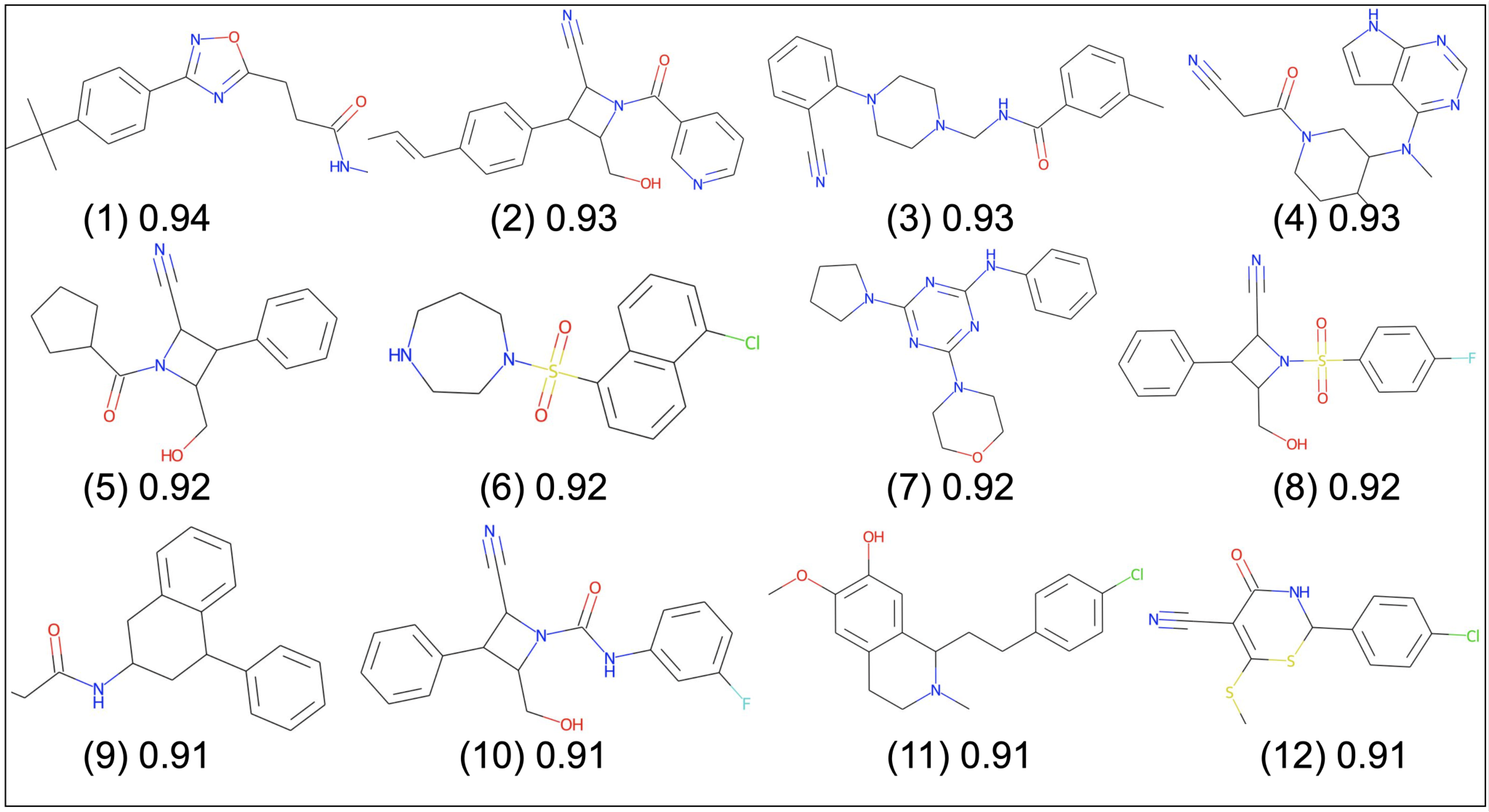}
\vspace{-10pt}
\caption{Top-12 molecular structures and QED scores for molecules in the training dataset.}
\label{fig:top_12_real}
\end{figure}
\begin{figure}[ht]
\centering
\includegraphics[width=0.7\hsize]{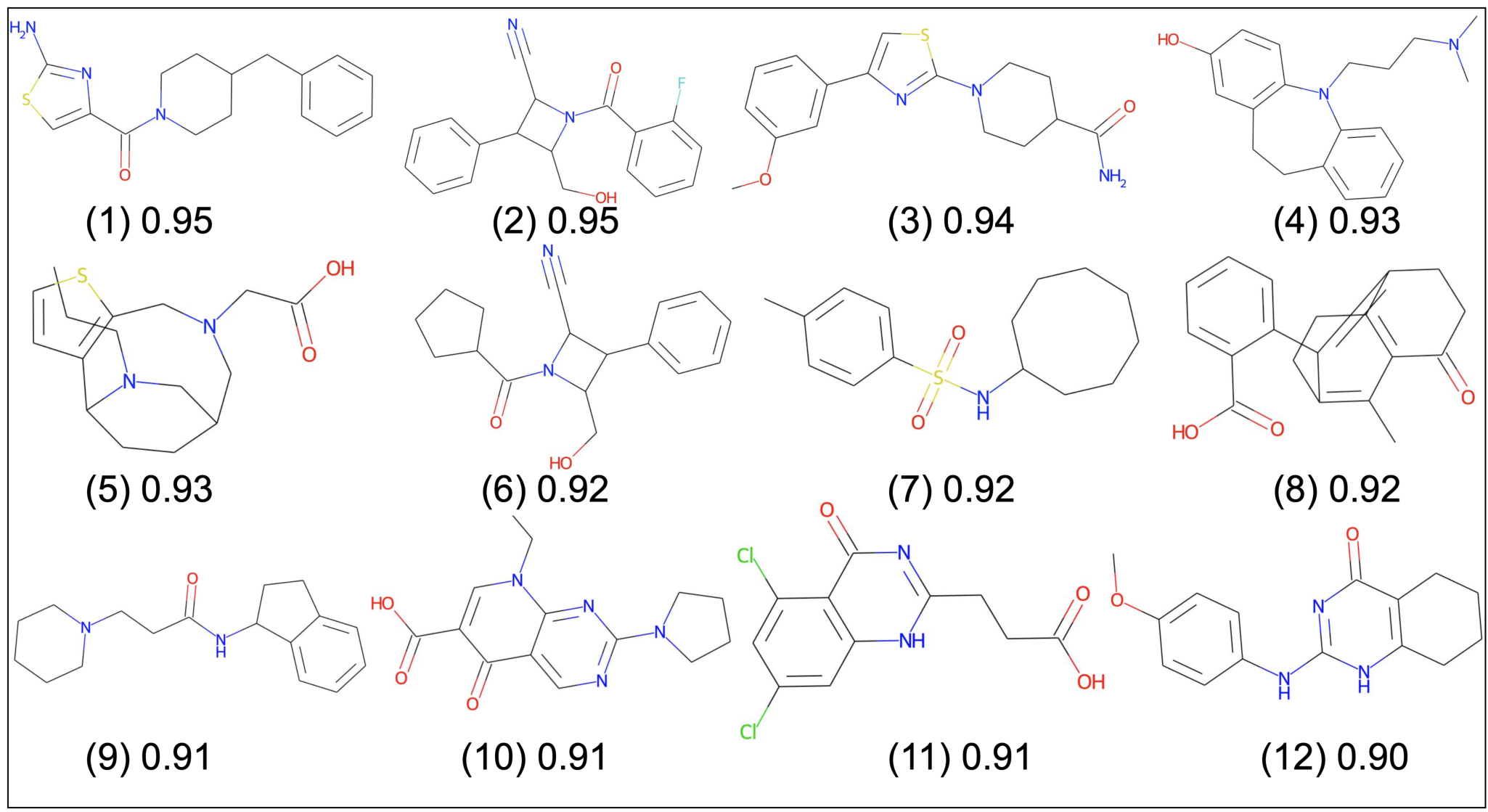}
\vspace{-10pt}
\caption{Top-12 molecular structures with QED scores for molecules generated by HVL2Mol.}
\label{fig:top_12_pred}
\end{figure}

Figures \ref{fig:top_12_real} and \ref{fig:top_12_pred} show the top-12 molecular structures with their QED scores for molecules in the training dataset and molecules generated by the proposed HVL2Mol model, respectively. It seems that all of the molecules are chemically valid and exhibit high QED scores. 

Furthermore, we evaluated the drug-likeness (QED) scores for the top-k generated molecules using the HVL2Mol model. The results are presented in Table \ref{table:top_k}. The molecules generated by HVL2Mol exhibited QED scores that were higher yet comparable to those of the training data. These findings demonstrate that the HVL2Mol model generated molecules while preserving the inherent QED properties.

\begin{table}[ht]
\setlength\tabcolsep{1pt}
\centering
\caption{Assessment of drug-likeness (QED) and synthesizability (SA) for the top-k generated molecules.}
\label{table:top_k}
\begin{tabular}{cccccc}\toprule
\makecell[c]{Chemical \\Property} & Data Source & Top-1 & Top-10 & Top-100 & Top-1000 \\\cmidrule{2-6}
\multirow{2}{*}{\makecell[c]{Drug-likeness \\(QED)}} & \makecell[c]{Compounds in \\training dataset} & 0.94 & 0.92 & \colorbox[gray]{0.9}{{\bf0.85}} & 0.64 \\ 
& \makecell[c]{Compounds generated \\by HVL2Mol} & \colorbox[gray]{0.9}{{\bf0.95}} & \colorbox[gray]{0.9}{{\bf0.93}} & 0.84 & \colorbox[gray]{0.9}{{\bf0.65}} \\\cmidrule{2-6}
\multirow{2}{*}{\makecell[c]{Synthesizability \\(SA)}} & \makecell[c]{Compounds in \\training dataset} & 1.00 & 0.94 & 0.85 & 0.47 \\ 
& \makecell[c]{Compounds generated \\by HVL2Mol} & \colorbox[gray]{0.9}{{\bf1.00}} & \colorbox[gray]{0.9}{{\bf0.99}} & \colorbox[gray]{0.9}{{\bf0.88}} & \colorbox[gray]{0.9}{{\bf0.48}} \\\toprule
\end{tabular}
\end{table}

\begin{figure}[t]
\centering
\includegraphics[width=0.7\hsize]{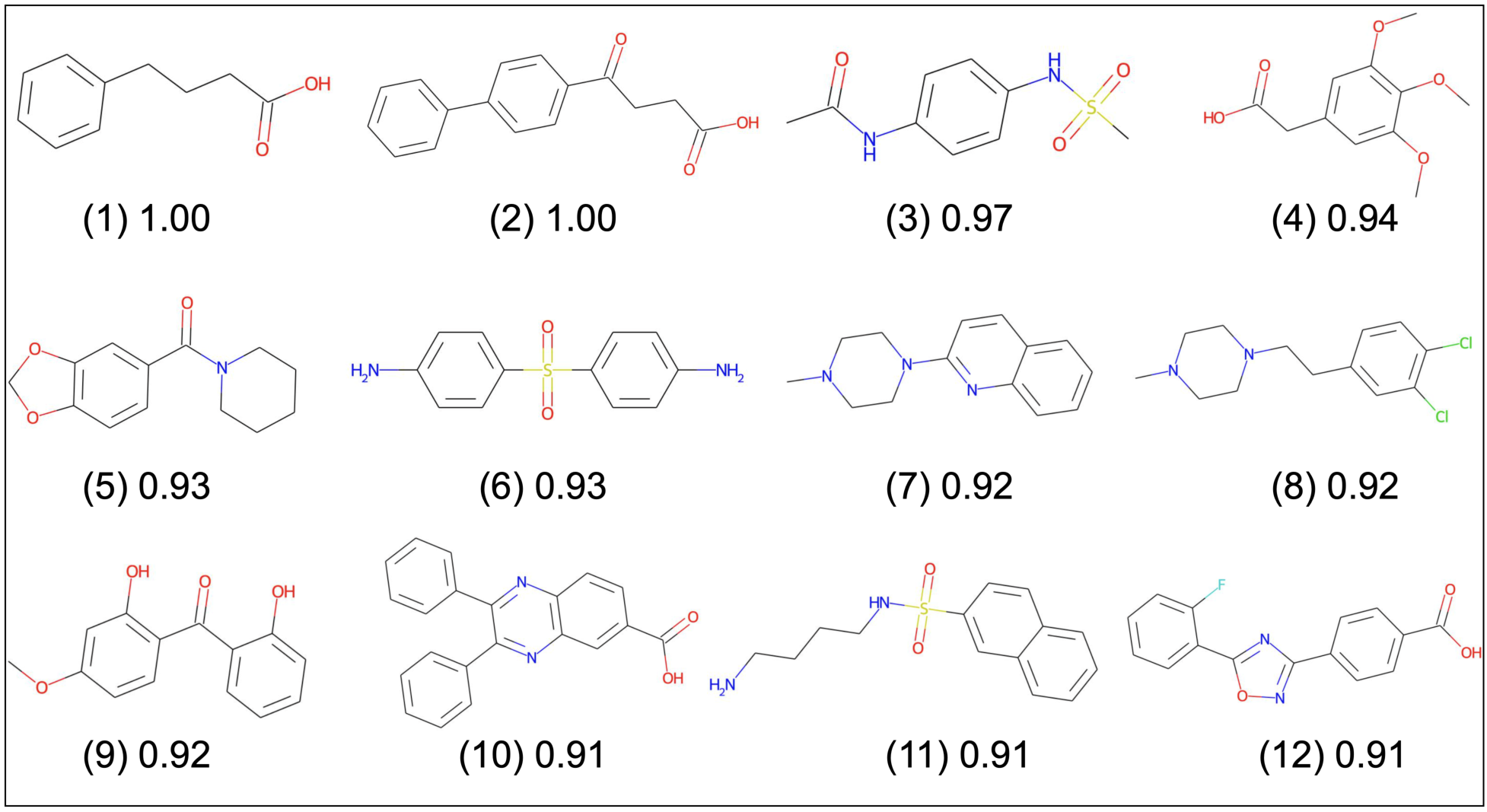}
\vspace{-10pt}
\caption{Top-12 molecular structures and their SA scores for molecules in the training dataset.}
\label{fig:top_12_real_sa}
\end{figure}

\begin{figure}[t]
\centering
\includegraphics[width=0.7\hsize]{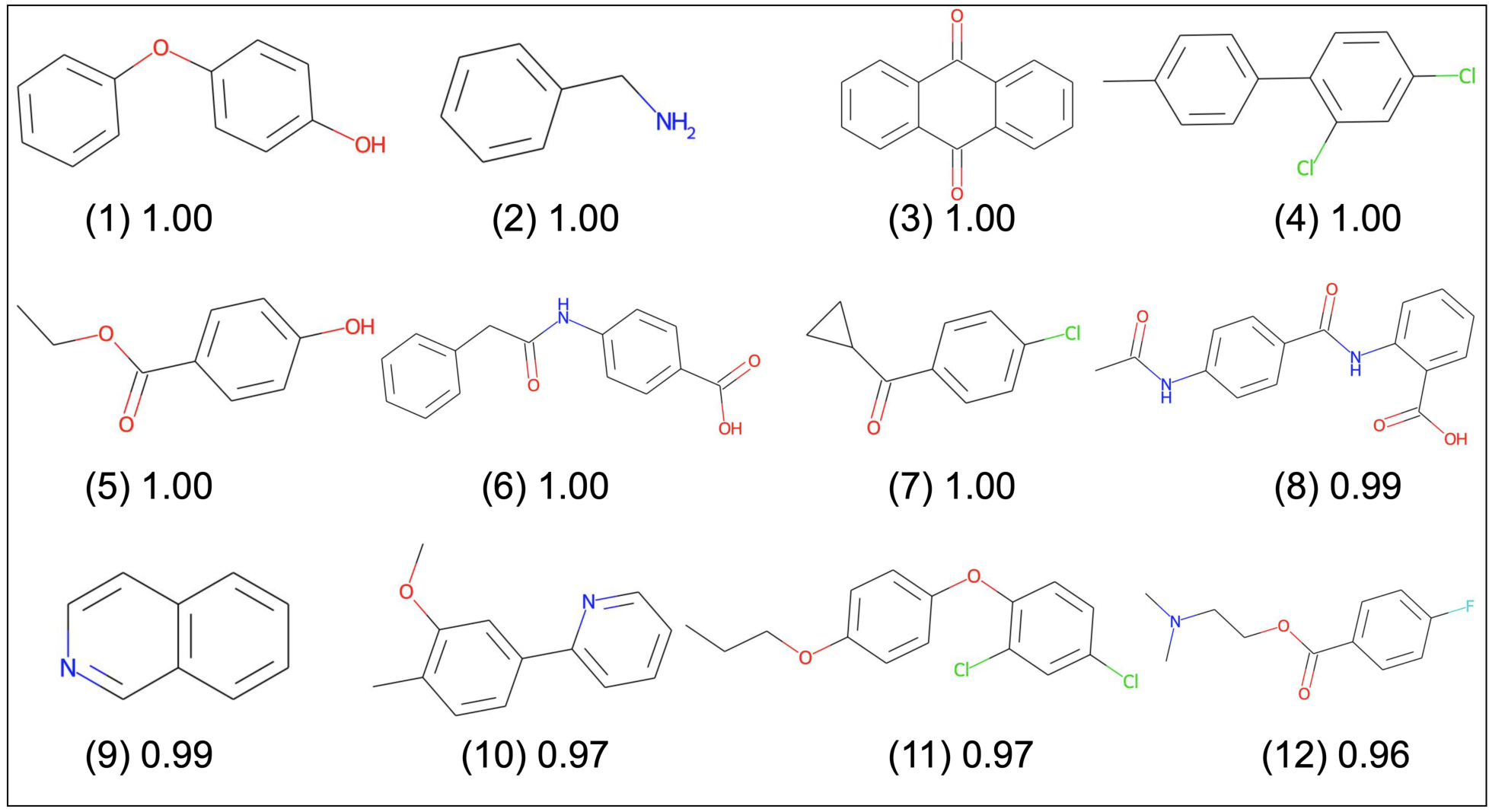}
\vspace{-10pt}
\caption{Top-12 molecular structures with their SA scores for molecules generated by HVL2Mol.}
\label{fig:top_12_pred_sa}
\end{figure}

Similarly, we present the top-12 molecular structures along with their synthesizability (SA) scores for molecules in the training dataset and those generated by HVL2Mol in Figures \ref{fig:top_12_real_sa} and \ref{fig:top_12_pred_sa}, respectively. The generated molecular structures indicate that our proposed HVL2Mol can produce valid molecules that are easy to synthesize. Moreover, the SA scores for the top-k generated molecules in Table \ref{table:top_k} demonstrate that HVL2Mol can effectively generate molecules with high SA scores. In other words, the molecules generated by HVL2Mol are confirmed to possess both drug-like characteristics and ease of synthesizability.

\subsection{HVL2Mol Generation}
\label{sec:generation}

\begin{figure*}[t]
\centering
\includegraphics[width=0.85\textwidth]{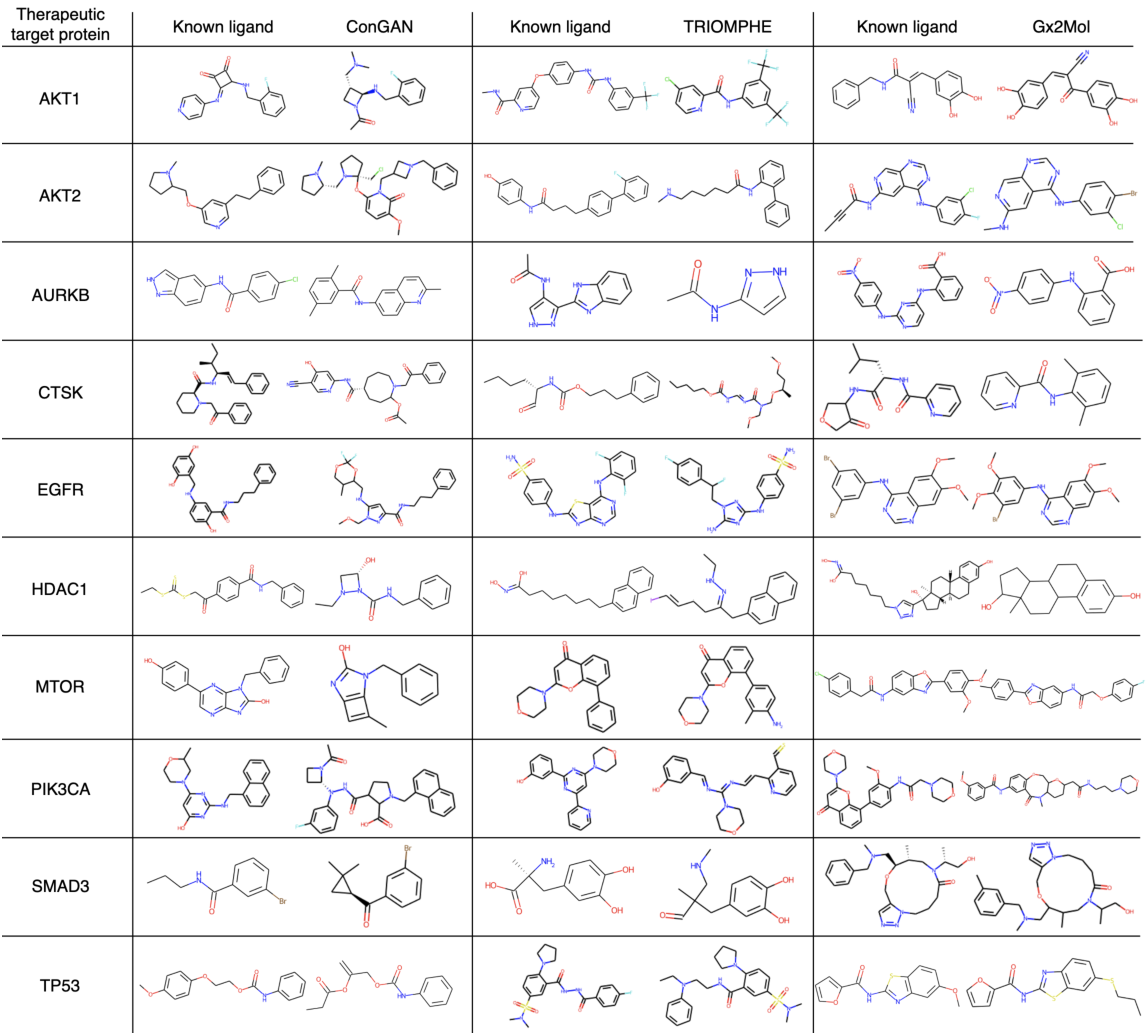}
\vspace{-10pt}
\caption{Comparison of newly generated molecules from the baseline and HVL2Mol. The 1st column of the table indicates the ten therapeutic target proteins. The 2nd, 4th, and 6th columns indicate known ligands for the corresponding target proteins. The 3rd, 5th, and 7th columns indicate newly generated molecules by ConGAN, TRIOMPHE, and HVL2Mol, which have the highest Tanimoto coefficients with the corresponding known ligands.}
\label{fig:molres}
\end{figure*}

Generally, the gene expression profiles of knockdown and overexpression of target proteins correlate with the gene expression profiles of inhibitors and activators, respectively \cite{kaitoh2021triomphe,narayanan2004single}. To generate molecules as candidates for ligands of target proteins, the gene expression profiles of the eight knockdown and two overexpressed target proteins were considered in this study. The former includes AKT1, AKT2, AURKB, CTSK, EGFR, HDAC1, MTOR, and PIK3CA. The latter includes SMAD3 and TP53. 

\begin{table}[t]
\setlength\tabcolsep{10pt}
\caption{Comparison of structural similarity scores between baseline and HVL2Mol methods.}
\label{tab:results}
\centering
\begin{threeparttable}
\begin{tabular}{lccc}
\hline
\makecell[c]{Therapeutic \\ target protein} & ConGAN & TRIOMPHE & HVL2Mol \\\hline
AKT1 & 0.32 & 0.42 & \colorbox[gray]{0.9}{{\bf0.53}} \\
AKT2 & 0.29 & 0.35 & \colorbox[gray]{0.9}{{\bf0.53}} \\
AURKB & 0.36 & 0.34 & \colorbox[gray]{0.9}{{\bf0.67}} \\
CTSK & 0.31 & 0.29 & \colorbox[gray]{0.9}{{\bf0.34}} \\
EGFR & 0.30 & 0.31 & \colorbox[gray]{0.9}{{\bf0.72}} \\
HDAC1 & 0.34 & 0.30 & \colorbox[gray]{0.9}{{\bf0.42}} \\
MTOR & 0.39 & \colorbox[gray]{0.9}{{\bf0.69}} & 0.46 \\
PIK3CA & 0.26 & \colorbox[gray]{0.9}{{\bf0.32}} & 0.30 \\\hline
SMAD3 & 0.44 & 0.48 & \colorbox[gray]{0.9}{{\bf0.85}} \\
TP53 & 0.46 & 0.53 & \colorbox[gray]{0.9}{{\bf0.55}} \\\hline
\end{tabular}
\begin{tablenotes}
\footnotesize
\item[$\star$] The values in bold in gray cells are the maximum values.
\end{tablenotes}
\end{threeparttable}
\end{table}

We conducted experiments on the newly generated molecules by comparing their molecular structures with those of the known ligands. If the newly generated molecules are meaningful, the newly generated molecules should be structurally similar to known ligands of each target protein to some extent. To ensure a fair comparison with the TRIOMPHE baseline, the default sampling number for each gene expression profile of the target protein was set to 1000, consistent with the setting used in TRIOMPHE. Subsequently, we only retained the valid molecules from the 1000 generated samples to calculate structural similarity using Tanimoto coefficients. The results are presented in Table \ref{tab:results}. ConGAN \cite{mendez2020novo} and TRIOMPHE \cite{kaitoh2021triomphe} are the two state-of-the-art (STOA) baseline models that are related to the proposed HVL2Mol model. For the former eight knockdown target proteins, six of the calculated Tanimoto coefficients for the molecules generated by the proposed HVL2Mol model with inhibitory ligands (i.e., AKT1, AKT2, AURKB, CTSK, EGFR, and HDAC1) outperformed the baseline methods. For MTOR and PIK3CCA, the Tanimoto coefficients performed second only to TRIOMPHE. In addition, for both 2SMAD3 and TP53, i.e., the target proteins with gene overexpression perturbations, the Tanimoto coefficients of the generated molecules by the proposed HVL2Mol model were higher than those obtained by the baseline methods. 

\begin{figure*}[t]
\centering
\includegraphics[width=0.8\textwidth]{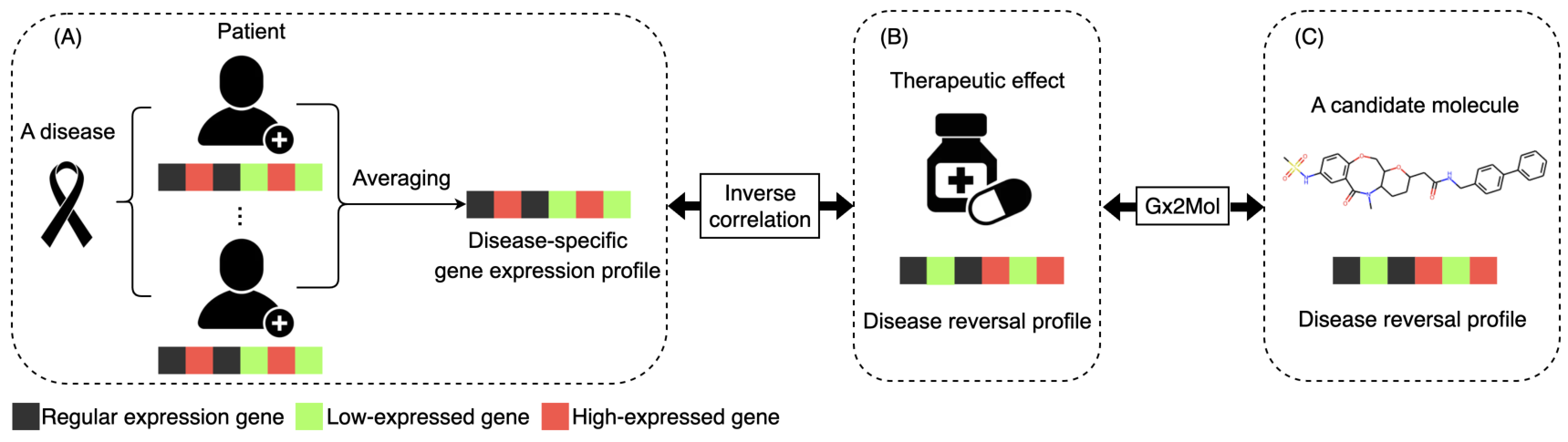}
\vspace{-10pt}
\caption{Data processing of gene expression profiles for the generation of therapeutic molecules.}
\label{fig:preprocessing}
\end{figure*}

Figure \ref{fig:molres} shows the molecules generated by the baseline and proposed HVL2Mol models. For the 10 target proteins, all generated molecules were structurally similar to the known ligands, compared with the baseline models. In summary, the proposed HVL2Mol model exhibited superior performance in terms of generating hit-like molecules from gene expression profiles via deep learning, and the proposed model outperformed the current SOTA baselines in most metrics.

\subsection{Case Studies}
\label{sec:case_study}
Generally, gene expression profiles are altered in a patient with a disease state. Accordingly, a molecule that counteracts the disease state is considered to have a therapeutic effect. Thus, in this case study, we attempted to generate molecules with therapeutic effects on a disease by considering disease-specific gene expression profiles. 

\begin{figure*}[t]
\centering
\includegraphics[width=0.8\hsize]{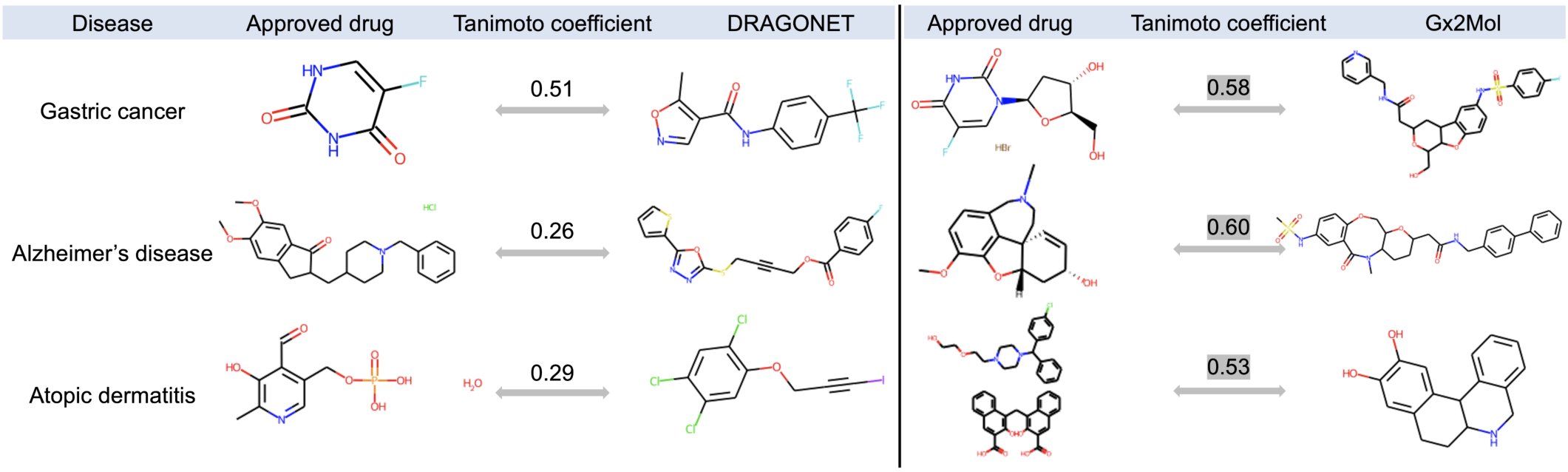}
\vspace{-10pt}
\caption{Assessing therapeutic impact: A comparison of Tanimoto coefficients and molecular structures between HVL2Mol and DRAGONET baseline model for newly generated molecules.}
\label{fig:disease}
\end{figure*}

Figure \ref{fig:preprocessing} illustrates the data processing of a gene expression profile for the generation of molecules with therapeutic effects on a disease. First, as shown in Figure \ref{fig:preprocessing} (A), a disease-specific gene expression profile is constructed by averaging the gene expression profiles of patients with a certain disease. Then, a gene expression profile that is inversely correlated with the disease-specific gene expression profile is constructed and defined as the disease reversal profile, as shown in Figure \ref{fig:preprocessing} (B). Finally, the disease reversal profile is used as an input to the proposed HVL2Mol model to generate molecules with therapeutic effects (Figure \ref{fig:preprocessing} (C)). 
The disease-specific gene expression profiles were obtained from the CREEDS database for patients with three diseases, i.e., gastric cancer, atopic dermatitis, and Alzheimer's disease. 

We examined the validity of the newly generated molecules by comparing the newly generated molecular structures with those of the approved drugs. If the newly generated molecules are meaningful, the newly generated molecules should be structurally similar to the approved drugs of each disease to some extent. We calculated the structural similarity using Tanimoto coefficients. Figure \ref{fig:disease} illustrates the Tanimoto coefficients between approved drugs and newly generated molecules, comparing the results obtained from the SOTA DRAGONET \cite{yamanaka2023novo} and our proposed HVL2Mol model, for each of the three diseases. Our proposed HVL2Mol model surpassed the SOTA DRAGONET in hit-like molecule generation for three diseases. HVL2Mol exhibited improved Tanimoto coefficients to approved drugs, reaching 0.58, 0.60, and 0.53 for gastric cancer, Alzheimer's disease, and atopic dermatitis. These improvements amounted to 13.7\%, 130.8\%, and 82.8\% for these three diseases. Additionally, fluorouracil (D04197) can be used in the treatment of liver metastases from gastrointestinal adenocarcinomas and also in the palliative treatment of liver and gastrointestinal cancers \cite{lu2014prediction}. When using the disease reversal profile of gastric cancer patients, the Tanimoto coefficient between the molecule generated by the proposed HVL2Mol model and fluorouracil was the largest. The Tanimoto coefficient of the HVL2Mol model generated molecule with floxuridine was maximum using the disease reversal profile of gastric cancer patients. These results suggest that the generated molecules effectively capture the structural features of approved anti-gastric cancer drugs. In addition, the molecules generated for the other two diseases demonstrate structural features that are similar to those of the approved drugs. As a result, the molecules generated using the proposed HVL2Mol model have potential drug-like properties. 

\section{Conclusion}
\label{sec:conclusion}
This study introduced the HVL2Mol model, designing to generate potential chemical structures of hit-like molecules from gene expression profiles using deep learning techniques. In the training phase, the HVL2Mol model first employed a VAE for feature extraction from high-dimensional gene expression profiles, and then the low-dimensional extracted features guided the generation of syntactically valid SMILES strings. In the generation phase, the VAE encoder served as the sole feature extractor, seamlessly combined with the generator to facilitate the generation of hit-like molecules. The results demonstrated the effectiveness of HVL2Mol in generating hit-like molecules from gene expression profiles. Additionally, a case study illustrates the model's ability to generate potential chemical structures for therapeutic drugs related to gastric cancer, stress dermatitis, and Alzheimer's disease using patients' disease reversal profiles.

This study has a primary limitation. Since LSTMs are frequently employed in auto-regressive generation tasks, wherein the token at the next time step is generated based on the token at the current time step, there exists a potential constraint on the diversity of generated molecules when using LSTMs as generators. In future research, we aim to explore strategies to enhance the diversity of molecular generation within the proposed HVL2Mol model. Furthermore, the envisaged application of the HVL2Mol model involves integration into practical AI systems to assist chemists in generating diverse drug candidate hit-like molecules tailored for various diseases. This integration is anticipated to leverage the strengths of the HVL2Mol model and contribute to the advancement of drug discovery processes.

\section*{Acknowledgements}
This research was supported by the International Research Fellow of Japan Society for the Promotion of Science (Postdoctoral Fellowships for Research in Japan [Standard]), AMED under Grant Number JP21nk0101111, and JSPS KAKENHI [grant number 20H05797].

\bibliographystyle{IEEEtran}
\bibliography{refs}

\begin{IEEEbiography}[{\includegraphics[width=1in,height=1.25in,clip,keepaspectratio]{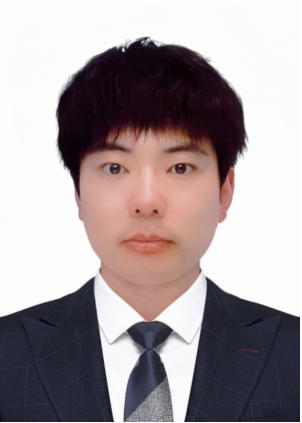}}]{Chen Li}
received the PhD degree in engineering from Hiroshima University, Japan, in 2019. In 2019 and 2020, he was a visiting researcher with the Graduate School of Advanced Science and Engineering, Hiroshima University, Japan. From 2021 to 2023, he was a research fellow with the Department of Bioscience and Bioinformatics, Faculty of Computer Science and Systems Engineering, Kyushu Institute of Technology, Japan. Since 2023, he has been a JSPS research fellow with the Graduate School of Informatics, Nagoya University, Japan. His research interests include deep learning and data mining, with research topics focusing on classification, prediction, and generation of time series data using AI techniques.
\end{IEEEbiography}

\begin{IEEEbiography}[{\includegraphics[width=1in,height=1.25in,clip,keepaspectratio]{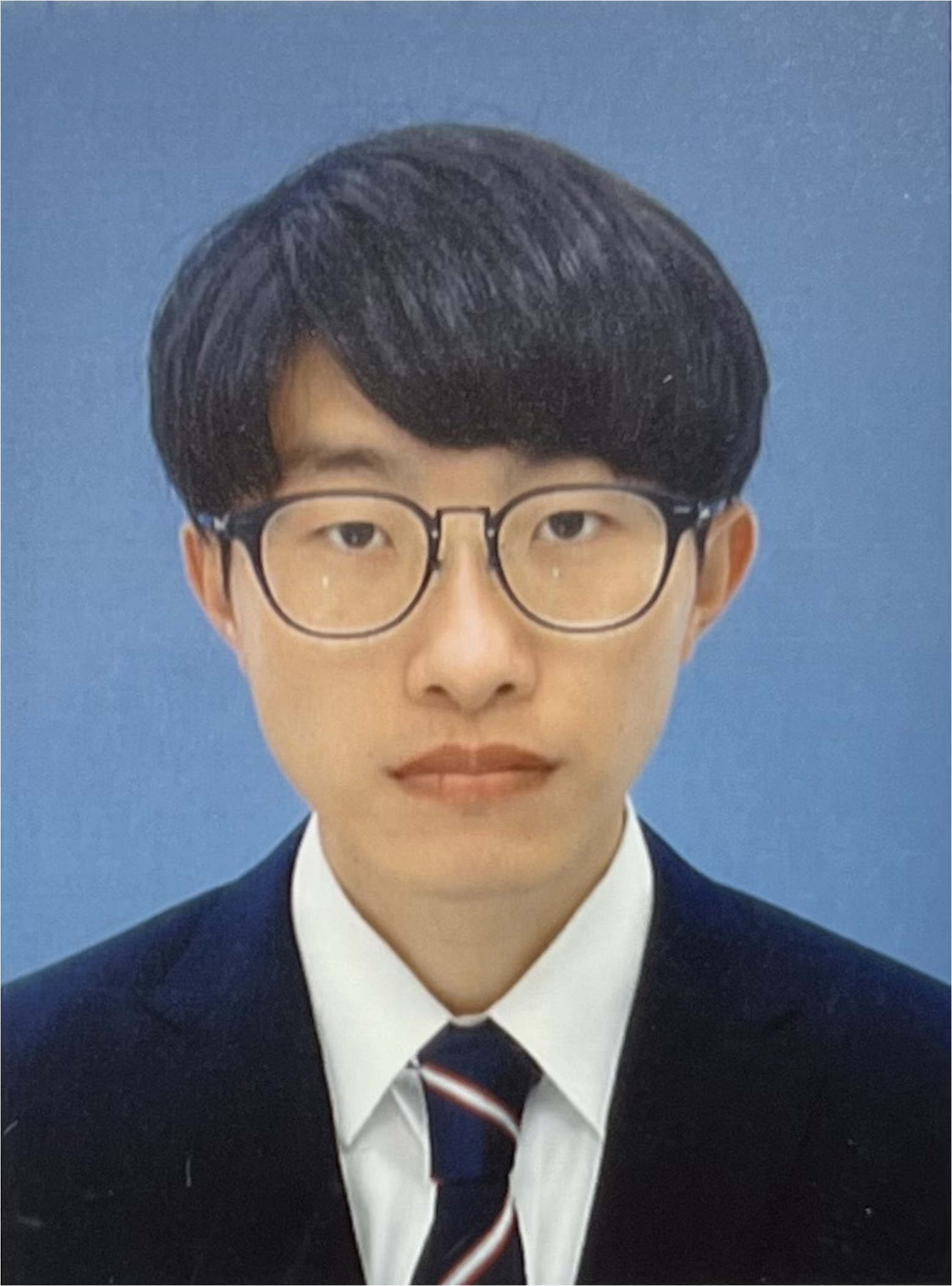}}]{Yuki Matsukiyo}
received the B.S. degree from the Department of Bioscience and Bioinformatics, Faculty of Computer Science and Systems Engineering, Kyushu Institute of Technology, Japan, in 2023. Since 2023, he has been a master’s student with the Department of Bioscience and Bioinformatics, Faculty of Computer Science and Systems Engineering, Kyushu Institute of Technology, Japan. His research interests focus on cheminformatics, molecular generation, and drug design.
\end{IEEEbiography}

\begin{IEEEbiography}[{\includegraphics[width=1in,height=1.25in,clip,keepaspectratio]{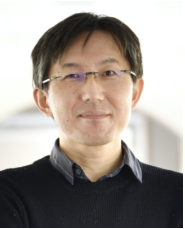}}]{Yoshihiro Yamanishi}
received the PhD degree from Kyoto University in 2005. He was a post-doctoral fellow at the Ecole des Mines de Paris from 2005 to 2006. He was assistant professor at Kyoto University from 2006 to 2007. He was a permanent researcher at Mines ParisTech and Curie Institute from 2008 to 2012. Since 2012, he was an associate professor (principal investigator) at the Medical Institute of Bioregulation, Kyushu University, Japan. From 2018 to 2023, he was a full professor with the Department of Bioscience and Bioinformatics, Kyushu Institute of Technology, Japan. Since 2023, he has been a full professor with the Department of Complex Systems Science, Graduate School of Informatics, Nagoya University, Japan. His research interest is in statistical machine learning methods for bioinformatics, chemoinformatics, and genomic drug discovery.
\end{IEEEbiography}

\end{document}